\documentclass[letterpaper, 10 pt, conference]{ieeeconf} 

\IEEEoverridecommandlockouts 
\usepackage{atbegshi,picture}
\usepackage{lipsum}
\usepackage{textcomp}
\usepackage{xcolor} 
\usepackage{graphicx} 

\newcommand{\gray}[1]{{\color{gray}\fontsize{10pt}{10pt}\selectfont #1}} 

\AtBeginShipout{\AtBeginShipoutUpperLeft{%
  \put(\dimexpr\paperwidth/2\relax,-1.25cm){\makebox[0pt][c]{
  \begin{tabular}{c}%
    \gray{This paper has been accepted for publication at the} \\ 
    \gray{IEEE International Conference on Robotics and Automation (ICRA), Atlanta, USA, 2025. \textcircled{c}IEEE} %
  \end{tabular}}%
  }}%
}

\usepackage{graphicx}
\usepackage{amsmath}
\usepackage{arydshln}
\usepackage{hhline}
\usepackage{eucal}
\usepackage{cite}
\usepackage{amssymb}
\usepackage{enumerate}
\usepackage[caption=false,font=footnotesize]{subfig}
\usepackage{bm}
\usepackage{color}
\usepackage{multirow}
\usepackage{algorithm2e}

\SetKwComment{Comment}{\scriptsize $\triangleright$\ }{}
\RestyleAlgo{ruled}

\usepackage{graphicx}
\usepackage{svg}
\usepackage{booktabs}
\usepackage{lipsum}
\usepackage{mathtools}

\newtheorem{rem}{Remark}

\providecommand{\norm}[1]{\lVert#1\rVert}

\makeatletter
\renewcommand{\env@matrix}[1][c]{%
  \hskip-\arraycolsep
  \let\@ifnextchar\new@ifnextchar
  \array{*\c@MaxMatrixCols #1}%
}
\makeatother

\DeclareSymbolFontAlphabet{\mathcal}   {symbols}

\usepackage{tikz}
\usetikzlibrary{calc}
\newcommand*\circled[1]{\tikz[baseline=(char.base)]{
    \node[shape=circle, draw, inner sep=1pt, 
        minimum height={\f@size*1.6},] (char) {\vphantom{WAH1g}#1};}}
\makeatother

\title{\LARGE \bf
CDM: Contact Diffusion Model for Multi-Contact Point Localization
}

\author{Seo Wook Han$^{1}$ and Min Jun Kim$^{1}$
\thanks{
This research has been funded 
by the Industrial Technology Innovation Program (P0028404) of the Ministry of Industry, Trade and Energy of Korea.
This work was supported 
by the National Research Foundation of Korea (NRF) grant funded by the Korea government (MSIT) No. 2021R1C1C1005232;
and by the Robot Industry Technology Development (RS-2024-00441872) funded By the Ministry of Trade, Industry \& Energy(MOTIE, Korea).
\textsuperscript{1}The authors are with Intelligent Robotic Systems Laboratory, Korea Advanced Institute of Science and Technology, Daejeon, Republic of Korea. {E-mail: {\tt\small first name.last name@kaist.ac.kr}}}
}
\begin{document}

\maketitle
\thispagestyle{empty}
\pagestyle{empty}

\begin{abstract}
In this paper, we propose a Contact Diffusion Model (CDM), a novel learning-based approach for multi-contact point localization. We consider a robot equipped with joint torque sensors and a force/torque sensor at the base. By leveraging a diffusion model, CDM addresses the singularity where multiple pairs of contact points and forces produce identical sensor measurements. We formulate CDM to be conditioned on past model outputs to account for the time-dependent characteristics of the multi-contact scenarios. Moreover, to effectively address the complex shape of the robot surfaces, we incorporate the signed distance field in the denoising process. Consequently, CDM can localize contacts at arbitrary locations with high accuracy. Simulation and real-world experiments demonstrate the effectiveness of the proposed method. In particular, CDM operates at 15.97ms and, in the real world, achieves an error of 0.44cm in single-contact scenarios and 1.24cm in dual-contact scenarios.
\end{abstract} 

\section{Introduction}
\label{sec:introduction}
Bringing robots into our daily lives presents many challenges.
In addition to efforts such as autonomous task execution \cite{10160294, chi2023diffusion}, another key focus is improving robots' \textit{physical interaction capabilities} to effectively handle unexpected collisions or contacts.
For example, \cite{10571557, 10611091, kim2019model, kim2019passivity, 9812268} propose compliant control methods that allow robots to safely interact with unknown environments or humans.
Other studies focus on rapidly detecting collisions for safety purposes \cite{kim2015design, heo2019collision, kim2021transferable, park2021collision}.
Several works have introduced contact localization methods to explicitly find contact points \cite{10161173, zwiener2019armcl, bimbo2019collision, popov2021multi, manuelli2016localizing}. 
Notably, \cite{10161173} and \cite{manuelli2016localizing} specifically address multi-contact scenarios.
Optimization-based control schemes have been presented to close the feedback loop using contact estimation results.
\cite{pang2022easing} addressed single-contact scenarios, while \cite{10611151, killpack2016model} addressed multi-contact scenarios.

\begin{figure}[t!]
\centering
    {\includegraphics[width=\columnwidth, trim = {0 0px 0 0px}, clip]{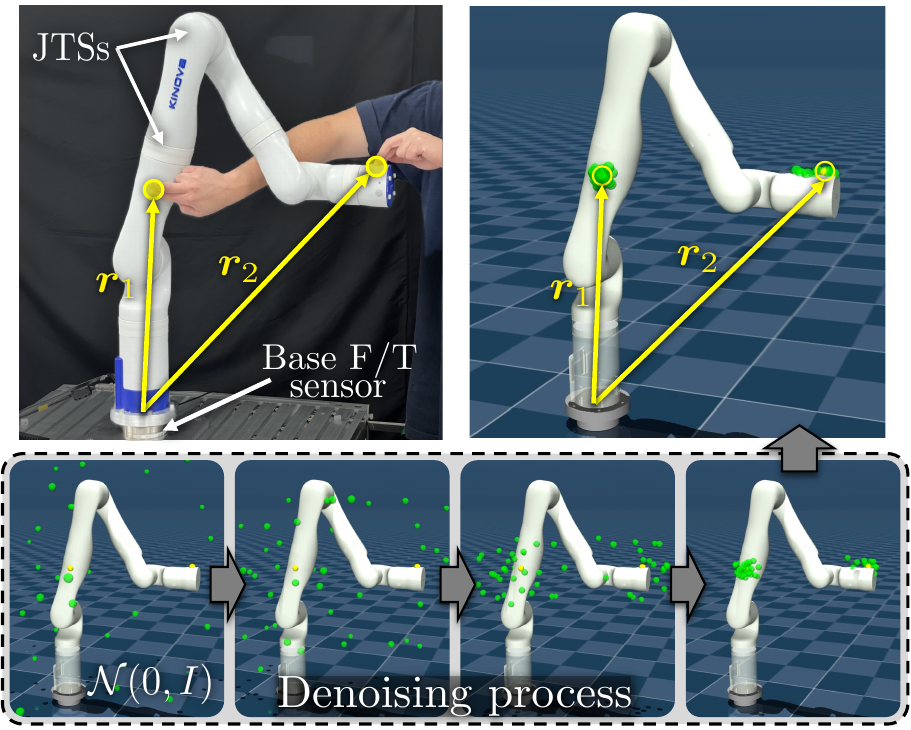}} 
    \caption{
    The objective of this paper is to localize the contact points ($\bm{r}_{i} \in \mathbb{R}^{3}$) by leveraging a diffusion model, where the denoising process is conditioned on proprioceptive sensor measurements, including JTSs and a base F/T sensor.
    The green colored points represent the samples generated by the proposed method.
    }
    \vspace{-7mm}
    \label{fig:introduction}
\end{figure}

Our research contributes to this area by proposing a method for \textit{multi-contact localization} (e.g., localizing $\bm{r}_{1}$ and $\bm{r}_{2}$ in Fig. \ref{fig:introduction}).
Similar to our previous work \cite{10161173}, we consider a collaborative robot equipped with proprioceptive sensors, specifically joint torque sensors (JTSs) and a force/torque (F/T) sensor at the base; see Fig. \ref{fig:introduction}.

\subsection{Related works}
Contact location estimation for \textit{single-contact} scenarios (i.e. there exists only one contact on the robot) has been well-studied.
Learning-based methods, such as those introduced in \cite{zwiener2018contact, popov2020transfer}, formulate the contact localization problem as a classification. 
In this approach, pre-labeled points on the robot are classified into contact/non-contact categories.
Additionally, in \cite{popov2020transfer}, the location of contact is parameterized as a value between 0 and 1 (from the base to the end-effector).
The aforementioned works simplified the robot surface; that is, they formulated the problem so that the solution corresponds to a point on the robot surface. 
This is due to (i) the kinematic relationships changing non-linearly with joint angles and (ii) the surface geometry being complex.

In contrast, model-based approaches, such as particle filter (PF)-based algorithms \cite{10161173, zwiener2019armcl, bimbo2019collision, popov2021multi, manuelli2016localizing}, handle surface constraints more effectively. 
Several methods have been proposed to keep particles representing the contact point constrained to the robot's surface.
For example, in \cite{10161173, zwiener2019armcl, bimbo2019collision, popov2021multi}, mesh data is used to represent the positions of particles, while in \cite{manuelli2016localizing}, particles are projected onto the robot surface.
With this setup, the particles are updated and resampled to find the contact location that best explains the sensor measurements, using the error metric defined via a quadratic programming (QP).

For both learning-based and PF-based approaches, singularity remains as a challenge. 
In contact localization problems, singularity occurs when multiple pairs of contact points and force vectors result in the same sensor measurements.
In this case, there exist multiple solutions for contact localization problems.
However, in PF-based algorithms, resampling process leads the particles to converge to a single point. 
Namely, PF-based algorithms output only one solution even if there are multiple solutions which have low QP errors.

There have been attempts to address the singularity problem. 
For example, \cite{pang2021identifying, popov2021real} proposed search algorithms for identifying multiple solutions.
In these methods, a set of contact points candidate is narrowed down across the entire robot using QP, but not in a greedy manner.
However, these methods are limited to \textit{single-contact} scenarios.

\textit{Multi-contact} scenarios present increased singularity, making them more difficult to handle.
To mitigate singularity, a key idea in PF-based algorithms is to use past observations.
Intuitively speaking, when dual-contact occurs sequentially and the location of the first contact is already known, the problem can be approached as solving a series of single contacts.
For instance, in \cite{10161173, manuelli2016localizing}, an additional particle set is initialized when the sensor measurements cannot be explained using a current set of particles.
However, initializing additional particle sets relies on a user-defined threshold that is difficult to tune.

It should be pointed out that the singularity is an inherent characteristic of the contact localization problem and cannot be entirely eliminated.
In this paper, therefore, our goal is not to eliminate singularity, but to predict it more accurately.
From a probabilistic perspective, the singularity problem suggests that the distribution of contact points is multi-modal, indicating uncertainty in localization.
Therefore, we aim to predict this multi-modal distribution more precisely.

\subsection{Contribution of the paper}
This paper proposes a method called Contact Diffusion Model (CDM) that tackles the aforementioned challenges:
(i) We address the uncertainty of localization by employing a diffusion model \cite{ho2020denoising, song2020score, luo2022understanding}, known for effectively handling multi-modal distributions.
To the best of our knowledge, this is the first attempt at contact localization using generative models.
(ii) Consequently, the proposed method does not require a difficult-to-tune threshold to distinguish between single-contact and dual-contact scenarios.
(iii) Inspired by PF-based algorithms, we modify the diffusion model to operate recursively using past model outputs.
This, in turn helps reduce multi-modality in cases where dual-contact occurs sequentially.
(iv) CDM does not rely on the simplified surface representations by employing signed distance field (SDF) of the robot surface.
Additionally, the inference time of CDM is $15.97\mathrm{ms}$.
In real-world validation, the accuracy of CDM is $0.44\mathrm{cm}$ for single-contact and $1.24\mathrm{cm}$ for dual-contact (for the single-contact case, \cite{zwiener2018contact} reports an error of $4\mathrm{cm}$ and \cite{popov2020transfer} shows an error of $6.4\mathrm{cm}$\footnote{To clarify, our setup involves additional sensing degrees of freedom at the base, enabling more precise localization, though this may complicate direct comparisons with the existing methods.}). 

\section{Preliminaries}
\label{sec:preliminaries}
Throughout this paper, we assume a point contact that does not include a moment \cite{10161173, zwiener2018contact, bimbo2019collision, manuelli2016localizing, popov2021multi, zwiener2019armcl, popov2020transfer}.

\subsection{Contact induced sensor measurements}
\label{subsec:contact_induced_sensor_measurements}
Let $\bm{r}_{i} \in \mathbb{R}^{3}$ and $\bm{F}_{i} \in \mathbb{R}^{3}$ be the $i^{\text{th}}$ contact point location and force, respectively. 
Suppose that $n_c$ contacts exist in the $n_q$-DOF robot arm (up to one contact per link), then the relation between the contacts and the available proprioceptive sensor measurements is given by
\begin{align}
    \bm{\mathcal{W}}_{ext} = \begin{bmatrix}
        \bm{\tau}_{ext} \\ \bm{\mathcal{F}}_{ext}
    \end{bmatrix}
    =\sum^{n_c}_{i=1}  \underbrace{\begin{bmatrix}
         \bm{J}_i^T(\bm{q}, \bm{r}_{i}) \\
        \bm{I}_{3\times3}  \\ 
        skew(\bm{r}_{i}) 
    \end{bmatrix}}_{\bm{A}_{i}(\bm{q},\bm{r}_{i})}
    \bm{F}_{i}, \label{eq:W_law}
\end{align}
where $\bm{\tau}_{ext}\in \mathbb{R}^{n_q}$ is external joint torque, 
$\bm{\mathcal{F}}_{ext} \in \mathbb{R}^{6}$ is the wrench at the base caused by contacts, 
$\bm{J}_{i}(\bm{q}, \bm{r}_{i}) \in \mathbb{R}^{3 \times n_q}$ is the associated positional Jacobian matrix at the contact point, 
$\bm{q}\in\mathbb{R}^{n_q}$ is the joint variable,
and $skew(\cdot)$ makes the $\mathbb{R}^3$ vector a $\mathbb{R}^{3 \times 3}$ skew-symmetric matrix. 
The estimated value of \eqref{eq:W_law}, denoted as $\hat{\bm{\mathcal{W}}}_{ext}$, can be obtained using a momentum-based disturbance observer (DOB) \cite{de2006collision,kim2015design} and an extended DOB \cite{buondonno2016combining, iskandar2021collision}.

\subsection{Optimization-based contact point localization}
\label{subsec:pf_based_contact_localization}
Using \eqref{eq:W_law}, we can formulate a convex optimization problem in which only the contact forces are decision variables:
\begin{equation}
\centering
\begin{gathered}
\label{eq:qp}
   QP(\hat{\bm{\mathcal{W}}}_{ext}\vert \bm{r}) = \displaystyle \min_{\bm{F}}  \left \| \hat{\bm{\mathcal{W}}}_{ext} - \sum_{i=1}^{n_c} \bm{A}_{i}(\bm{q},\bm{r}_{i})  \bm{F}_{i}   \right \|^{2} \\
    \text{subject to} \,\,\, \bm{F}_{i} \in \mathcal{F}_c(\bm{r}_{i}), \;\;  \forall i \in \{1,\ldots,n_c\},
\end{gathered}
\end{equation}
where $\bm{r} = \{\bm{r}_{i}\}^{n_c}_{i=1}$, $\bm{F} = \{\bm{F}_{i}\}^{n_c}_{i=1}$, 
and $\mathcal{F}_c(\bm{r}_{i})$ represents the friction cone at $\bm{r}_{i}$ to mathematically express the force in the pushing direction only.
The QP \eqref{eq:qp} can be used as a likelihood, i.e., we can evaluate the contact point candidates by computing \eqref{eq:qp} to determine how well the contact points can explain the sensor measurements $\hat{\bm{\mathcal{W}}}_{ext}$.
For example, in PF-based algorithms \cite{manuelli2016localizing, zwiener2019armcl, bimbo2019collision, 10161173, popov2021multi}, \eqref{eq:qp} is used to determine the weight of particles.
Although \eqref{eq:qp} is not used in the proposed method, it is employed in Section~\ref{subsec:exp1} to evaluate the points generated by CDM.

\subsection{Denosing diffusion probabilistic models (DDPM)}
\label{subsec:ddpm}
The goal of diffusion models \cite{ho2020denoising, song2020score, luo2022understanding} is to generate a new sample $\hat{x}$ from the data distribution $p(x)$ by training on the dataset $x^{[0]}\sim p(x)$.
Diffusion models involve two primary processes: the forward and backward process (also known as diffusion and denoising process).
In the forward process, the scheduled noise is sequentially added to the given dataset by running a Markovian forward diffusion process:
\begin{align}
q(x^{[k]} \vert x^{[k-1]}) = {\mathcal{N}}(x^{[k]} ; \sqrt{1-\beta_k}x^{[k-1]}, \beta_k {I}),
\end{align}
where $k = 1, \hdots, K$ represents the diffusion step, $K$ is the number of diffusion steps, and $\beta_k$ is the noise scale at $k$. 
The number of diffusion steps is large enough that $x^{[K]}$ becomes Gaussian white noise, i.e., $x^{[K]} \sim {\mathcal{N}}(0, {I})$.
With this setup, the data distribution at $k$ can be obtained in closed form:
\begin{align}
\label{eq:reparameterized_q}
q(x^{[k]} \vert x^{[0]}) = {\mathcal{N}}(x^{[k]} ; \sqrt{\bar{\alpha}_k}x^{[0]}, (1-\bar{\alpha}_k) I),
\end{align}
where $\alpha_k = 1-\beta_k$ and $\bar{\alpha}_k = \prod_{i=1}^k \alpha_i$.

The reverse process transforms $x^{[K]}$ into a sample that follows the data distribution by sequentially eliminating the noise using the following learned Gaussian distribution \cite{ho2020denoising}:
\begin{align}
\label{eq:unconditional_backward_process}
p_{\theta}(x^{[k-1]} \vert x^{[k]}, k) = {\mathcal{N}}(x^{[k-1]} ; {\mu}_{\theta}(x^{[k]}, k), {\tilde{\beta}}_k {I}),
\end{align}
where $\theta$ is the learnable network parameter, ${\mu}_{\theta}(x^{[k]}, k) = \frac{1}{\sqrt{\alpha_k}} \left( x^{[k]} - \frac{1-\alpha_k}{\sqrt{1-\bar{\alpha}_k}} {\epsilon}_{\theta}(x^{[k]}, k) \right)$,  ${\epsilon}_{\theta}(x^{[k]}, k)$ is the output of the denoiser network, and $\tilde{\beta}_k = \beta_k (1- \bar{\alpha}_{k-1})/(1-\bar{\alpha}_k)$.
The denoiser network, ${\epsilon}_{\theta}(x^{[k]}, k)$, is trained via a simplified loss function \cite{ho2020denoising}:
\begin{align}
\label{eq:unconditional_loss}
{\mathcal{L}}_\theta = {\mathbb{E}}_{k,x^{[0]},{\epsilon}} \norm{{\epsilon} - {\epsilon}_{\theta}(x^{[k]}, k)}^2,
\end{align}
where $k \sim \mathcal{U}(1,K)$, which is the uniform distribution, $x^{[k]}$ is the sample from \eqref{eq:reparameterized_q}, and ${\epsilon} \sim {\mathcal{N}}(0,{I})$.

\section{Contact Diffusion Model (CDM)}
\label{sec:mcdm}
\subsection{Problem statements}
To account for the time dependency in multi-contact point localization problem, we define $\bm{r}_{t} = \{\bm{r}_{t,i} \}^{n_c}_{i=1} \in \mathbb{R}^{n_c \times 3}$ as $n_c$ contact points at time $t$.
Then, multi-contact point localization problem can be formulated as estimating the following posterior:
\begin{align}
\label{eq:posterior_basic}
    p(\bm{r}_{t} \vert \bm{O}_{1:t}, \mathcal{S}),
\end{align}
which is the probability distribution over the contact points $\bm{r}_{t}$, conditioned on all past observations ($\bm{O}_{1:t}$), and the robot surfaces $\mathcal{S}$.
Since the contact points should be on the robot surfaces, \eqref{eq:posterior_basic} is conditioned on $\mathcal{S}$.
However, this posterior cannot be directly addressed by a diffusion model, because the observation time length cannot be infinite.

To effectively address the posterior \eqref{eq:posterior_basic} with a diffusion model, we consider cases with up to two sequential contacts ($n_c \leq 2$). 
When the last single contact occurs at time $T_s$, the posterior representing the multi-contact localization problem is expressed as follows:
\begin{align}
\label{eq:posterior_assump}
    p(\bm{r}_{t} \vert \bm{O}_{t-T:t}, {\bm{r}}_{T_s},\mathcal{S}),
\end{align}
where $\bm{O}_{t-T:t}$ is the time sequence observations from $t-T$ to $t$, and $T$ is the observation time length.

\subsection{Dataset of CDM}
Let us define a contact dataset $\mathcal{D} = \{ \xi_i \}^{N_d}_{i=1}$, where $\xi$ contains the location of contact points and the corresponding time sequence observations.
We use multiple samples regardless of the number of contacts $n_c$.
Specifically, $\xi = \{ \bm{X}_t^{[0]}, \bm{O}_{t-T:t} \}$, where $\bm{X}_t \in \mathbb{R}^{n_p \times 3}$ is a duplication of $\bm{r}_t$, and $n_p$ is the number of points.
The reasons for this are as follows:
(i) The dimension of $\bm{r}_{t}$ changes depending on the number of contacts $n_c$, making it difficult to handle both single and dual-contact cases with a single model.
(ii) With $n_c$ sampled points, it is challenging to capture multi-modality; e.g., consider the case when $n_c = 2$ and the posterior \eqref{eq:posterior_assump} has three modes.

For effective training, the observation consists of 
(i) the available proprioceptive sensor measurements $\hat{\bm{\mathcal{W}}}_{ext}$, 
(ii) the joint angles $\bm{q}$, and 
(iii) the position and orientation of all the links, i.e., $\bm{T}_l = (\bm{p}_l, \textbf{q}_{l}), \forall l \in \{1, \hdots, n_q \}$, where $\bm{p}_l \in \mathbb{R}^{3}$ and $\textbf{q}_l \in \mathbb{R}^{4}$ represent the position and quaternion of the $l^{\text{th}}$ link frame, respectively.

\begin{figure}[t!]
\centering
    {\includegraphics[width=0.97\columnwidth, trim = {0 0px 0 0px}, clip]{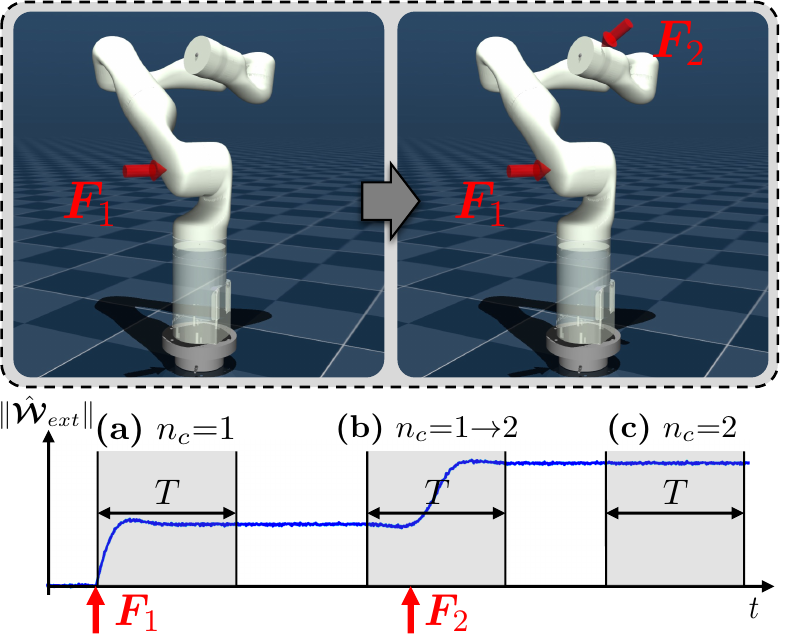}}
    \caption{
    The dataset is collected from scenarios where dual-contact occurs sequentially.
    We refer to the three types of observations as (a) single-contact, (b) transition dual-contact, and (c) steady dual-contact.
    }
    \vspace{-6mm}
    \label{fig:dataset}
\end{figure}

As shown in Fig. \ref{fig:dataset}, we collect the dataset in scenarios where dual-contact occurs sequentially.
Depending on the number of contacts within the observation time window, we refer to the contact state of $\xi$ as single-contact ($n_c=1$), transition dual-contact ($n_c=1$ $\rightarrow$ $2$), or steady dual-contact ($n_c=2$).
This provides comprehensive coverage of possible contact scenarios for training and evaluation.

\begin{figure*}[t!]
\centering
	{\includegraphics[width=\textwidth, trim = {0 0px 0 0px}, clip]{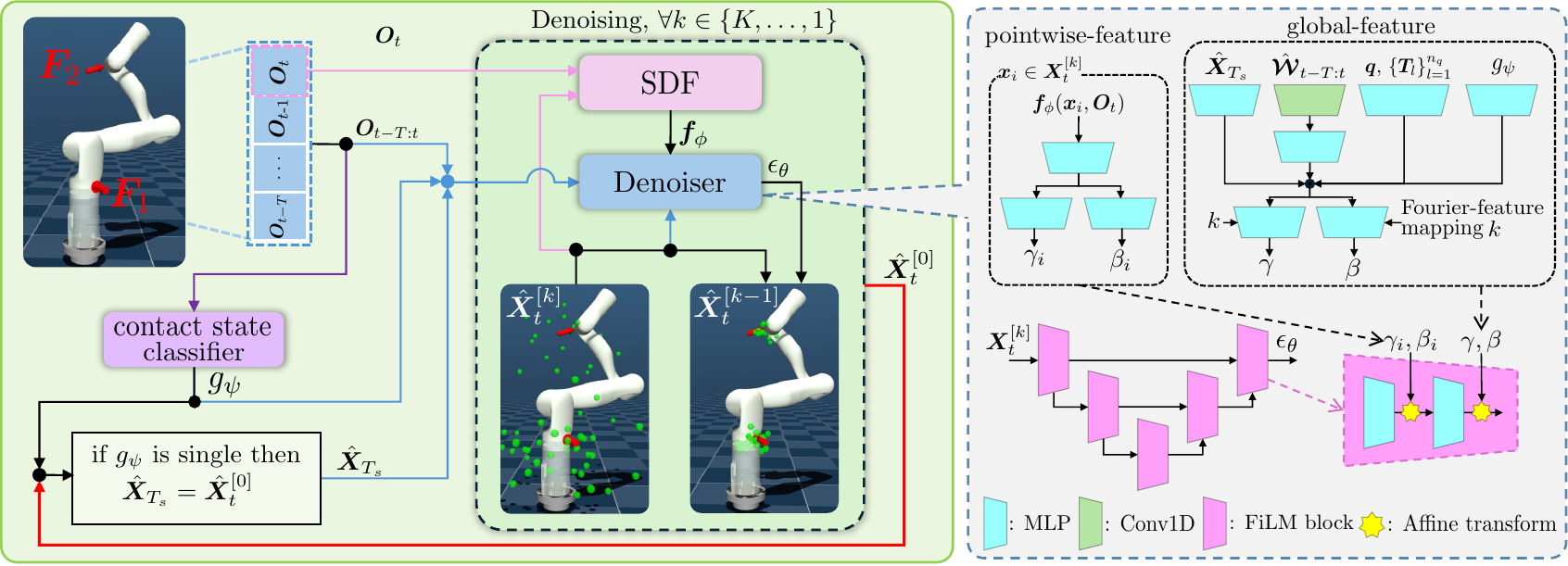}}
	\caption{
    In the inner loop, the denoiser sequentially eliminates noise from the white Gaussian noise (from $k=K$ to $1$). 
    To incorporate the surface constraints on the points, we utilize the pre-trained SDF network for each link surface.
    In the outer loop, the contact state classifier determines whether to use the diffusion output from the previous time step.
    On the right, the structure of the denoiser network is shown.
	}
 \vspace{-6mm}
	\label{fig:network_structure}
\end{figure*}
\subsection{Architecture of CDM}
Let us rewrite the posterior \eqref{eq:posterior_assump} using the notation $\bm{X}_t$:
\begin{align}
\label{eq:posterior}
    p(\bm{X}_{t} \vert \bm{O}_{t-T:t}, {\bm{X}}_{T_s},\mathcal{S}).
\end{align}
Then, the goal of CDM is to generate samples from the posterior \eqref{eq:posterior}, i.e., $\hat{\bm{X}}_{t} \sim p_\theta(\bm{X}_{t} \vert \bm{O}_{t-T:t}, \hat{{\bm{X}}}_{T_s},\mathcal{S})$.
The architecture of the proposed method is illustrated in Fig. \ref{fig:network_structure}.
In the remainder of this subsection, we present how the posterior \eqref{eq:posterior}, conditioned on $\bm{X}_{T_s}$ and $\mathcal{S}$, can be approximated by the diffusion model.

\subsubsection{Historical diffusion result conditioning}
The training process of CDM is presented in Algorithm~\ref{alg:mcdm_train}.
Our key idea is that $\hat{{\bm{X}}}_{T_s}$ is sampled from the first contact point in the dataset $\bm{r}_{t,1}^{[0]}$.
Specifically, we apply the diffusion process \eqref{eq:reparameterized_q} to the first contact point, using a randomly sampled diffusion step $k_h \sim \mathcal{U}(1,K)$.
Intuitively, in the dual-contact case, the denoiser network is trained to find the contact points, whether or not the first contact point is given (i.e., whether the sampled $k_h$ is close to 1 or not).

Algorithm~\ref{alg:mcdm_inference} presents the inference process of CDM.
To iteratively update the $\hat{{\bm{X}}}_{T_s}$ only in single-contact cases, the contact state classifier $g_{\psi}$ is utilized, where $\psi$ is a learnable parameter.
The details of $g_{\psi}$ will be presented in the next subsection.

\begin{rem}
    The proposed training process allows us null-conditioning, i.e., $\hat{\bm{X}}_{t} \sim p_\theta(\bm{X}_{t} \vert \bm{O}_{t-T:t}, \emptyset, \mathcal{S})$ by setting $\hat{\bm{X}}_{T_s} \sim \mathcal{N}(0, I)$.
    Because the denoiser network is also trained to handle cases where $k_h$ is set close to $K$.
    In other words, CDM can operate even in situations where there is no information about the first contact. 
    For instance, this occurs when Algorithm~\ref{alg:mcdm_inference} starts for the first time or when two contacts are simultaneously applied to the robot.
\end{rem}

\begin{algorithm}
\DontPrintSemicolon
\SetKwInput{Input}{Input}
\SetKwInput{Output}{Output}
\caption{CDM-Training}\label{alg:mcdm_train}
\SetAlgoLined
\Input{ Training dataset $\mathcal{D}$, denoiser network $\epsilon_\theta$, learning rate $\alpha$
}
\While{\textit{training is not finished}}{
    \Comment*[l]{\scriptsize sample a batch of contact points}
    $\{\bm{X}_t^{[0]}, \bm{O}_{t-T:t}\} \sim \mathcal{D}$, $\epsilon \sim \mathcal{N}(0,I)$, $k \sim \mathcal{U}(1,K)$ \;
    $\bm{X}_t^{[k]} =  \sqrt{\bar{\alpha}_k}\bm{X}_t^{[0]} + \sqrt{1-\bar{\alpha}_k} \epsilon$ \;
    \Comment*[l]{\scriptsize run forward process to the first contact point}
    $\epsilon_h \sim \mathcal{N}(0,I)$, $k_h \sim \mathcal{U}(1,K)$ \;
    $\hat{\bm{X}}_{T_s} = \sqrt{\bar{\alpha}_{k_h}}\bm{r}_{t,1}^{[0]} + \sqrt{1-\bar{\alpha}_{k_h}} \epsilon_h$ \;
    \Comment*[l]{\scriptsize compute the loss function}
    $\mathcal{L}_\theta = \norm{\epsilon - \epsilon_\theta(\bm{X}_t^{[k]}, \bm{O}_{t-T:t}, \hat{\bm{X}}_{T_s},  k)} $ \; 
    \Comment*[l]{\scriptsize update parameters}
    $\theta = \theta + \alpha \nabla_{\theta} \mathcal{L}_{\theta}$ \;
    }
\end{algorithm}

\begin{algorithm}
\DontPrintSemicolon
\SetKwInput{Input}{Input}
\SetKwInput{Output}{Output}
\caption{CDM-Inference}\label{alg:mcdm_inference}
\SetAlgoLined
\Input{Trained noise prediction network $\epsilon_\theta$, trained contact state classifier $g_{\psi}$
}
$\hat{\bm{X}}_{T_s} \sim \mathcal{N}(0,I)$\;
\While{\textit{robot is in contact}}{
\Comment*[l]{\scriptsize get the observations from the robot}
$\bm{O}_{t-T:t}$ = \textit{get\_observations}\;
\Comment*[l]{\scriptsize start from the Gaussian white noise}
 $\hat{\bm{X}}_{t}^{[K]}  \sim \mathcal{N}(0,I)$\;
\Comment*[l]{\scriptsize sequentially eliminate the noise}
\For{$k = K,\ldots,1$}{
    $\epsilon_\theta = \epsilon_\theta(\hat{\bm{X}}_t^{[k]}, \bm{O}_{t-T:t}, \hat{\bm{X}}_{T_s}, k)$ \;
    $\hat{\bm{X}}_t^{[k-1]} \sim {\mathcal{N}}(\frac{1}{\sqrt{\alpha_k}} \left( \hat{\bm{X}}_t^{[k]} - \frac{1-\alpha_k}{\sqrt{1-\bar{\alpha}_k}} {\epsilon}_{\theta} \right), {\tilde{\beta}}_k {I})$ 
}
\Comment*[l]{\scriptsize utilization of historical result}
\If{$g_{\psi}(\bm{O}_{t-T:t})$ \textit{is single-contact}}
{
    $\hat{\bm{X}}_{T_s} = \hat{\bm{X}}_{t}^{[0]}$
}
}
\end{algorithm}

\subsubsection{Surface constraints conditioning}
To guide the generated samples to be located on the robot surfaces, the denoiser network is conditioned on the SDF of all link surfaces, $\bm{f}_{\phi}$ (the detailed conditioning structure will be presented in the next subsection).
This is obtained by applying the SDF network to each link surface.
Specifically, for the link $l$ and a point $\bm{x}_i \in \bm{X}_t^{[k]}$, the SDF network output is defined as
\begin{align}
\label{eq:sdf}
    ({d}_l, \bm{g}_l) = f_{\phi_l}(\bm{x}_i, \bm{O}_t) \in \mathbb{R}^{4},
\end{align}
where ${d}_l \in \mathbb{R}^{+}$ is the shortest distance from point $\bm{x}_i$ to the link $l$, $\bm{g}_l \in \mathbb{R}^{3}$ is a unit vector toward the link $l$, $\phi_l$ is the learnable parameter for the SDF network of the link $l$, and $\bm{O}_{t}$ is the current observation.
As a result, the SDF outputs for all link surfaces are represented as $\bm{f}_{\phi}(\bm{x}_i, \bm{O}_t) = \{f_{\phi_l}(\bm{x}_i, \bm{O}_t)\}_{l=1}^{n_q}$.
In our implementation, we adopt the method from \cite{rebain2021deep} to compute \eqref{eq:sdf}.

\subsection{Neural network model}
\textbf{Denoiser:} The structure of the denoiser is shown in Fig.~\ref{fig:network_structure}.
The overall structure is similar to U-Net \cite{ronneberger2015u} where the convolution layers are replaced by multi-layer perceptrons (MLPs).
The denoiser network incorporates a conditioning structure using FiLM \cite{perez2018film} and Fourier-feature mapping \cite{tancik2020fourier} to generate results based on $\bm{O}_{t-T:t}$, $\hat{\bm{X}}_{T_s}$, pre-trained network outputs ($\bm{f}_{\phi}$ and $g_{\psi}$), and $k$.
As shown in Fig.~\ref{fig:network_structure}, for each FiLM block, we utilize the two FiLM layers: one for pointwise feature and the other for global feature conditioning.
The first FiLM layer, which takes $\bm{f}_{\phi}(\bm{x}_i, \bm{O}_t)$ as input, generates outputs for each individual point, i.e., $\bm{x}_i \leftarrow \gamma_i \bm{x}_i + \beta_i$ for all $\bm{x}_i \in \bm{X}_t^{[k]}$. 
In contrast, the second FiLM layer applies a common $\gamma$ and $\beta$ to all points, i.e., $\bm{X}_t^{[k]} \leftarrow \gamma \bm{X}_t^{[k]} + \beta$.
Additionally, excluding the estimated external values $\hat{\bm{\mathcal{W}}}_{ext}$, we only use the current values of the kinematic-related observations.

\textbf{Contact state classifier:} 
We slightly modify the second FiLM layer of the denoiser so that its output becomes a one-hot label of $n_c$. 
The loss function is then defined as cross-entropy loss:
\begin{align}
    \mathcal{L}_\psi = {\mathbb{E}}_{y, \bm{O}_{t-T:t}} \left( y \log(g_{\psi}) + (1 - y) \log(1 - g_{\psi}) \right),
\end{align}
where $y \sim \mathcal{D}$ is one-hot label of $n_c$, and $\bm{O}_{t-T:t} \sim \mathcal{D}$.

\section{Experimental Validation}
\label{sec:simulation_validation}
The proposed method, CDM, is implemented with JAX \cite{jax2018github}, and the QP \eqref{eq:qp} is solved with qpSWIFT \cite{pandala2019qpswift}.
Training and evaluation were performed on a system with an Intel Core i7-12700 CPU, an RTX 3090 GPU, and 64GB of RAM, using Python and C++ programs.
All experiments are conducted with a 7-DOF Kinova Gen3 robot arm equipped with JTSs and an F/T sensor (ATI Mini85) at the base.

\noindent\textbf{Dataset:} 
The dataset was collected with MuJoCo \cite{6386109}.
A robot with a random configuration was simulated, operating at $1000\mathrm{Hz}$ for $300\mathrm{ms}$. 
Two contact forces were sequentially applied to the robot at $0\mathrm{ms}$ and $150\mathrm{ms}$. 
Each contact force, lying within the friction cone, had a magnitude randomly selected between $10$ and $25\mathrm{N}$.
Additionally, for more realistic data, noise is added to the JTSs and base F/T sensor measurements.
Out of 6,144K simulated trajectories, 90\% were used for training and 10\% for evaluation. 
Each trajectory was divided into segments using a random stride ranging from 15 to 45 steps, with an observation time length of $T=60\mathrm{ms}$, and each segment was used as a data point $\xi \in \mathcal{D}$.
Moreover, we set the number of samples, $n_p=64$,  to balance inference speed and the ability to capture the multi-modality.

\noindent\textbf{Implementation and training details:}
We employ a diffusion model with $K=1,000$, using linearly scheduled noise from $\beta_1 = 10^{-6}$ to $\beta_K = 10^{-3}$. 
The denoiser, with 143M parameters, was trained for 259K steps with a batch size of 1,024 for approximately 24 hours. 
The contact state classifier, which has 15M parameters, was trained for 5K steps with a batch size of 1,024 for approximately 1 hour.
In the following experiments, we use 10 denoising steps with DDIM \cite{song2020denoising}.

\noindent\textbf{Inference time:}
The average inference time of CDM is $15.97\mathrm{ms}$ (time for 10 denoiser inferences). 
While CDM is slower than our previous work \cite{10161173}, PF-based algorithms require time for particle convergence. 
Taking this into account, our previous work took $8.21\mathrm{ms}$ for single-contact and $18.26\mathrm{ms}$ for dual-contact. 
Note that CDM requires only one step to localize contacts and maintains consistent speed regardless of the number of contacts.

\begin{figure}[tbh!]
\centering
    {\includegraphics[width=\columnwidth, trim = {0 0px 0 0px}, clip]{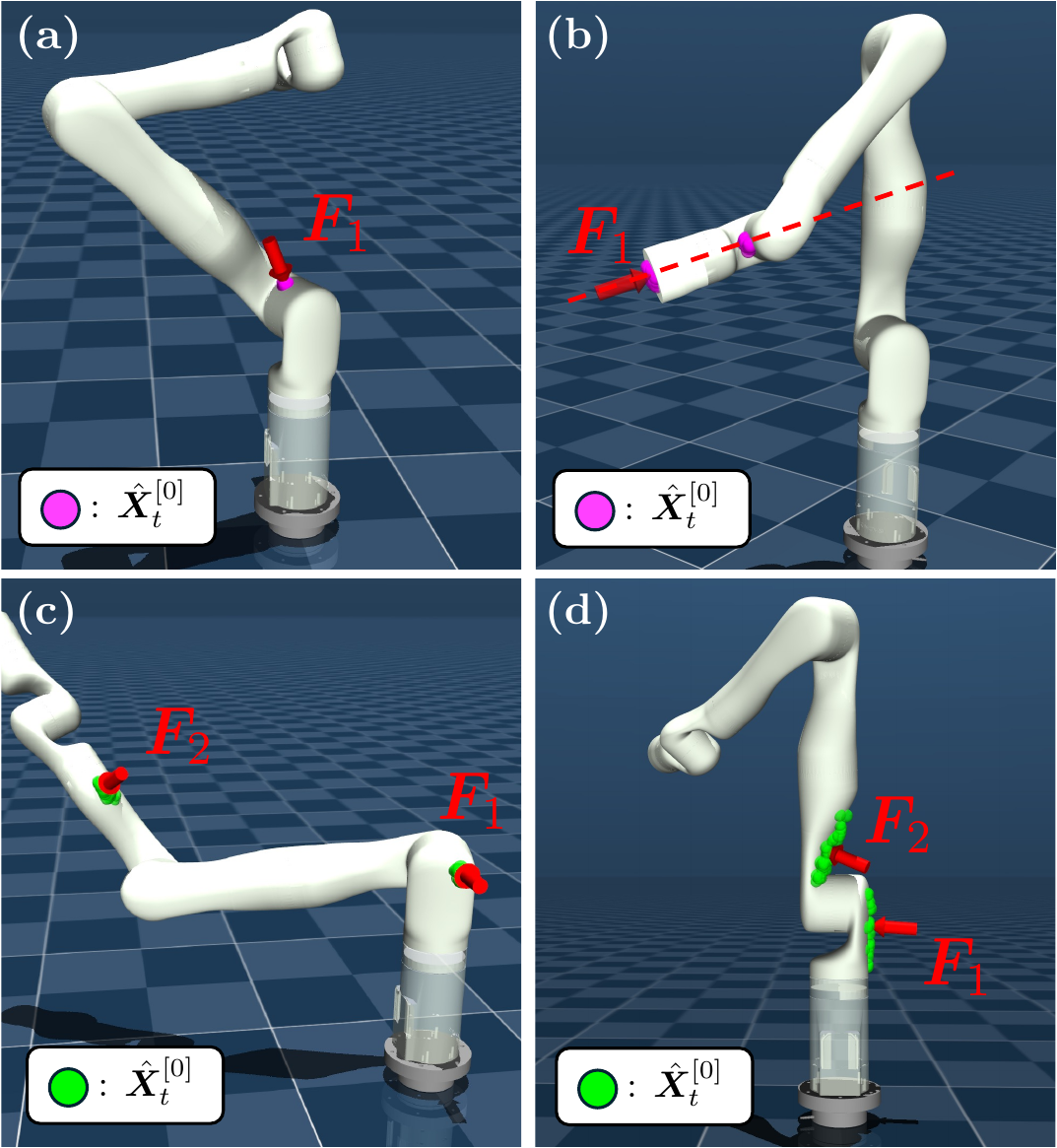}}
    \caption{
    \textbf{Experiment \#1:} The generated samples from CDM, corresponding to different contact states:
    (a) Single-contact.
    (b) Single-contact, but the posterior \eqref{eq:posterior} has two modes.
    (c) Transition dual-contact.
    (d) Steady dual-contact.
    }
    \vspace{-6mm}
    \label{fig:exp1_multi_modality}
\end{figure}

\subsection{Evaluating the feasibility of generated samples}
\label{subsec:exp1}
In experiment \#1, we aim to validate feasibility of the samples generated by CDM.
Since there is no historical results in the evaluation dataset, we employ the null-conditioning, i.e., $\hat{\bm{X}}_{t} \sim p_{\theta}(\bm{X}_t \vert \bm{O}_{t-T:t}, \emptyset, \mathcal{S})$.
For quantitative evaluation of CDM, we report the following metrics:
\textbf{(i) Failure rate: }
We assess the contact state classifier by measuring the failure rate for the contact state.
\textbf{(ii) M-RMSE:=} $\frac{1}{n_c} \sum_{j=1}^{n_c} \min_{\bm{x}_i \in \hat{\bm{X}}_{t}}\norm{\bm{r}_{t,j}^{[0]} - \bm{x}_i}$.
Small M-RMSE means that the generated samples $\hat{\bm{X}}_{t}$ contain the contact points in dataset $\bm{r}_{t}^{[0]}$.
\textbf{(iii) QP error: } 
This metric is defined as average value of $\norm{\hat{\bm{\mathcal{W}}}_{ext} - {\bm{\mathcal{W}}}^*}$, where $\bm{\mathcal{W}}^*$ is the optimal solution of the QP \eqref{eq:qp} when the contact points candidates are given by $\hat{\bm{X}}_{t}$.
A small QP error indicates that $\hat{\bm{X}}_{t}$ explains the observation well.
Moreover, we report three separate metrics of QP error to provide a clearer physical interpretation\footnote{$\bm{\mathcal{W}}^*$ is the concatenation of values induced from the JTSs, base force, and base torque sensors.}-\textit{JTSs error} $(\mathrm{N\cdot m})$, \textit{Base F error} $(\mathrm{N})$, and \textit{Base T error} $(\mathrm{N\cdot m})$. 
For instance, \textit{JTSs error} is defined as $\norm{\hat{\bm{\tau}}_{ext} - \bm{\tau}_{ext}^*}$.

\begin{table}[htb!]
\resizebox{\columnwidth}{!}{%
\begin{tabular}{cccccccc}
\hline
\begin{tabular}[c]{@{}c@{}} contact \\ state \end{tabular} & \begin{tabular}[c]{@{}c@{}}\# of \\ data \end{tabular} & \begin{tabular}[c]{@{}c@{}} M- \\ RMSE ($\mathrm{cm}$) \end{tabular} & \begin{tabular}[c]{@{}c@{}} Failure\\ rate (\%)\end{tabular} & \begin{tabular}[c]{@{}c@{}}QP\\ error \end{tabular} & \begin{tabular}[c]{@{}c@{}} JTSs \\ error ($\mathrm{N \cdot m}$)\end{tabular} & \begin{tabular}[c]{@{}c@{}}Base F \\ error ($\mathrm{N}$) \end{tabular} & \begin{tabular}[c]{@{}c@{}}Base T \\ error ($\mathrm{N \cdot m}$) \end{tabular}  \\ \hhline{========}
Single	    &2,395,243	&0.32	&0.15	&0.39	&0.29	&0.12	&0.20	\\ \hline
Trans-Dual	&1,016,760	&0.76	&0.00	&0.59	&0.44	&0.18	&0.30	\\ \hline
Steady-Dual	&2,089,949	&1.14	&1.83	&0.99	&0.70	&0.33	&0.49	\\ \hline
\end{tabular}%
}
\caption{Experiment \#1 results }
\label{tab:exp1_overview}
\vspace{-6mm}
\end{table}

The results are summarized in Table~\ref{tab:exp1_overview}, and examples of generated samples are presented in Fig. \ref{fig:exp1_multi_modality}.
For single-contact cases, as shown in Fig. \ref{fig:exp1_multi_modality}(a), CDM can accurately localize the single-contact point.
While this occurs in most cases, CDM can also generate samples for both modes when the posterior \eqref{eq:posterior} exhibits multi-modality in a singular case (see Fig. \ref{fig:exp1_multi_modality}(b)).
Moreover, we observed that, in most cases, transition dual-contact exhibits less multi-modality compared to steady dual-contact;
the samples shown in Fig. \ref{fig:exp1_multi_modality}(c) and \ref{fig:exp1_multi_modality}(d) all have a QP error of less than 0.1, but the latter shows more spread out results.

\subsection{Ablation study on SDF conditioning}
\label{subsec:exp_sdf}
In experiment \#2, we demonstrate the effectiveness of the SDF-embedded denoiser network by comparing CDM with and without SDF. 
In the case without SDF, there is no FiLM layer for pointwise feature conditioning (see Fig. \ref{fig:network_structure}).
When SDF is included, the average distance between the generated samples and the robot surfaces is $0.29\mathrm{cm}$, whereas without SDF, it is $0.89\mathrm{cm}$.

\subsection{Ablation study on historical diffusion result conditioning}
\label{subsec:exp2}
\begin{figure}[tbh!]
\centering
    {\includegraphics[width=0.95\columnwidth, trim = {0 0px 0 0px}, clip]{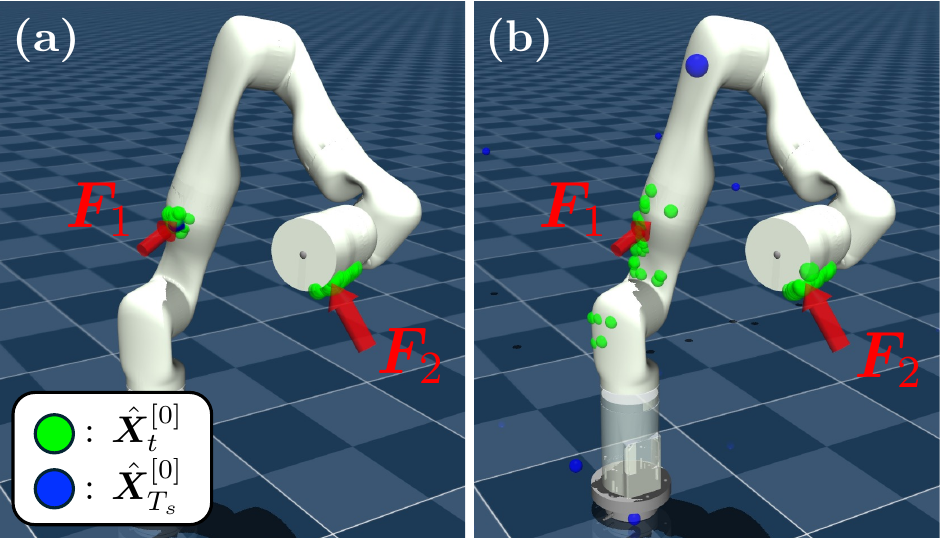}}
    \caption{
    \textbf{Experiment \#3}:
    The samples generated by CDM: (a) with historical result conditioning, and (b) without it.
    }
    \vspace{-6mm}
    \label{fig:exp2_explaination}
\end{figure}

In experiment \#3, we demonstrate that historical result conditioning helps reduce the multi-modality of \eqref{eq:posterior}.
For simplicity, we denote \textit{CDM-his} as the CDM that uses historical diffusion results, and \textit{CDM-null} as the CDM that does not.
The trajectory simulated over $300\mathrm{ms}$ is divided according to the inference time of CDM.
For example, the input sequence for \textit{CDM-his} is $\{\bm{O}_{0:60}, \bm{O}_{16:76},\hdots,\bm{O}_{240:300}\}$ with updated $\hat{\bm{X}}_{T_s}$, based on Algorithm~\ref{alg:mcdm_inference}.
For \textit{CDM-null}, only the last observation, $\bm{O}_{240:300}$, is provided as input, with $\hat{\bm{X}}_{T_s} \sim \mathcal{N}(0,I)$.
As shown in Fig. \ref{fig:exp2_explaination}, relying solely on observations increases multi-modality and spreads the points (even though the generated samples explain the sensor measurements well), whereas incorporating single-contact information reduces multi-modality. 
To make a quantitative comparison, the following metrics are reported, including M-RMSE:
\textbf{(i) C-RMSE-$\bm{j}$:=}
$\norm{\bm{r}_{t,j}^{[0]}-\bm{c}_j}$, where $\bm{c}_j \in \mathbb{R}^{3}$ is the center of the $j^{\text{th}}$ k-means cluster of $\hat{\bm{X}}_{t}^{[0]}$, and the number of clusters is equal to the number of contacts $n_c$.

\begin{table}[htb!]
\centering
\resizebox{\columnwidth}{!}
{%
\begin{tabular}{ccccc}
\hline
                            & \# of data & M-RMSE ($\mathrm{cm}$) & {C-RMSE-1 ($\mathrm{cm}$)} & {C-RMSE-2 ($\mathrm{cm}$)} \\ \hhline{=====}
\textbf{\textit{CDM-his}}       & 614,400 &0.62  & 1.63                   & 2.53     \\ \hline
\textbf{\textit{CDM-null}}      & 614,400 &1.33  & 3.89                   & 6.09     \\ \hline
\begin{tabular}[c]{@{}c@{}}\textbf{PF-based} \\ \textbf{algorithm}\cite{10161173}\end{tabular}  & 10,000  &1.08  & N/A                   & N/A     \\ \hline
\end{tabular}
}
\caption{Experiment \#3 results}
\vspace{-6mm}
\label{tab:exp2}
\end{table}

We conducted this experiment 614.4K times ($10\%$ of the total dataset before it was divided).
The results, including those from our previous work \cite{10161173}, are presented in Table~\ref{tab:exp2}.
For all metrics (M-RMSE and C-RMSE), \textit{CDM-his} shows more than twice the improvement compared to \textit{CDM-null}.
Moreover, the proposed method shows about a 40\% performance improvement over our previous work \cite{10161173}, which is a PF-based algorithm.
Strictly speaking, the metric used in PF-based algorithms \cite{10161173} differs from M-RMSE, as it evaluates the localization error between a single solution and the contact points in the dataset.
However, since CDM is designed to estimate a multi-modal distribution rather than to provide a single solution, this comparison would be the most appropriate.
Additionally, this experiment was conducted using realistic observations (including sensor noise and accounting for the convergence time of the DOB; see Fig. \ref{fig:dataset}), presenting a more challenging setup compared to our previous work.
As will be presented shortly, in experiment \#4, we demonstrate that the model trained on the simulation dataset performs well in real-world scenarios.

\subsection{Direct sim-to-real transfer}
\label{subsec:real_world}
In experiment \#4, we validate CDM using a real-world robot (see Fig. \ref{fig:introduction} and the attached video, which is also available at \cite{youtube}).
To obtain the locations of the true contact points, we attached labeled markers to the robot.
Similar to experiment \#3, for the dual-contact case, the contact forces were sequentially applied to the robot.
Single and dual contact scenarios were conducted 100 times each, and every 20 experiments, the robot configuration was randomly set.

\begin{table}[tbh!]
\resizebox{\columnwidth}{!}
{
\begin{tabular}{ccccc}
\hline
contact state & \# of data  & M-RMSE ($\mathrm{cm}$) & C-RMSE-1 ($\mathrm{cm}$) & C-RMSE-2 ($\mathrm{cm}$) \\ \hhline{=====}
single        & 100        & 0.44      & 0.68         & N/A            \\ \hline
dual        & 100          & 1.24       & 2.61         & 2.85         \\ \hline
\end{tabular}
}
\caption{Experiment \#4 results}
\vspace{-6mm}
\label{tab:exp_real}
\end{table}

Table~\ref{tab:exp_real} summarizes the real-world experiment results.
In the single-contact case, the proposed method achieves higher accuracy than existing learning-based methods (with \cite{zwiener2018contact} and \cite{popov2020transfer} reporting errors of $4\mathrm{cm}$ and $6.4\mathrm{cm}$, respectively).
To the best of our knowledge, this is the first quantitative evaluation of dual-contact scenarios in the real-world.

\section{Conclusion}
\label{sec:conclusion}
In this work, we propose a method called CDM (Contact Diffusion Model), to solve multi-contact point localization problem.
To the best of authors' knowledge, this is the first attempt to localize contact using generative models.
By exploring characteristics of the multi-contact localization problem, we modify the posterior to be conditioned on past model outputs.
This allows us to estimate the posterior with reduced multi-modality when dual-contact occurs sequentially.
Moreover, in the denoising process, we employ the SDF-embedded denoiser network to guide the samples to be on the robot's surfaces.
In the validation, we demonstrate that CDM effectively captures the multi-modality of the posterior.
We also presented ablation studies to show the effectiveness of SDF and historical result conditioning.
Additionally, we conducted quantitative evaluations including the real-world experiments for both single and dual-contact scenarios.
As a future work, we will extend CDM to include contact force identification, enabling simultaneous estimation of contact point locations and forces.

\bibliographystyle{myIEEEtran.bst}
\bibliography{IEEEabrv,bib_ICRA2025_mcdm.bib}

\end{document}